\documentclass{article}

\PassOptionsToPackage{numbers, compress}{natbib}
\usepackage{array}
\usepackage{makecell}
\usepackage[final]{neurips_data_2023}
\usepackage{eqnarray}
\usepackage{caption}
\usepackage{subcaption}

\usepackage{color}
\usepackage{caption} 

\usepackage[utf8]{inputenc} 
\usepackage[T1]{fontenc}    
\usepackage{hyperref}       
\usepackage{url}            
\usepackage{booktabs}       
\usepackage{amsfonts}       
\usepackage{nicefrac}       
\usepackage{microtype}      
\usepackage{xcolor}         

\usepackage{adjustbox}
\usepackage{multirow}
\usepackage{tabularx}
\usepackage{array}
\usepackage{graphicx}
\usepackage{wrapfig}

\usepackage{caption}

\usepackage{amsmath}
\usepackage{bm}

\usepackage{xcolor}
\usepackage{hyperref}

\hypersetup{
    colorlinks=true,
    citecolor=green,
    linkcolor=red,
    urlcolor=magenta,
}

\title{Mask2Map: Vectorized HD Map Construction Using Bird's Eye View Segmentation Masks}

\author{
  Sehwan Choi$^1$\thanks{Equal contribution.} \quad Jungho Kim$^1$\footnotemark[1] \quad Hongjae Shin$^1$ \quad \textbf{Jun Won Choi$^2$\thanks{Corresponding author.}}
\\
  $^1$ Hanyang University \hspace{0.5cm}
  $^2$ Seoul National University\\
}

\begin{document}

\maketitle

\begin{abstract}
In this paper, we introduce Mask2Map, a novel end-to-end online HD map construction method designed for autonomous driving applications. Our approach focuses on predicting the class and ordered point set of map instances within a scene, represented in the bird's eye view (BEV).
Mask2Map consists of two primary components: the \textit{Instance-Level Mask Prediction Network} (IMPNet) and the \textit{Mask-Driven Map Prediction Network} (MMPNet). IMPNet generates Mask-Aware Queries and BEV Segmentation Masks to capture comprehensive semantic information globally. Subsequently, MMPNet enhances these query features using local contextual information through two submodules: the \textit{Positional Query Generator} (PQG) and the \textit{Geometric Feature Extractor} (GFE). PQG extracts instance-level positional queries by embedding BEV positional information into Mask-Aware Queries, while GFE utilizes BEV Segmentation Masks to generate point-level geometric features.
However, we observed limited performance in Mask2Map due to inter-network inconsistency stemming from different predictions to Ground Truth (GT) matching between IMPNet and MMPNet. 
To tackle this challenge, we propose the \textit{Inter-network Denoising Training} method, which guides the model to denoise the output affected by both noisy GT queries and perturbed GT Segmentation Masks.
Our evaluation conducted on nuScenes and Argoverse2 benchmarks demonstrates that Mask2Map achieves remarkable performance improvements over previous state-of-the-art methods, with gains of 10.1\% $mAP$ and 4.1\% $mAP$, respectively. Our code can be found at \href{https://github.com/SehwanChoi0307/Mask2Map}{https://github.com/SehwanChoi0307/Mask2Map}

\end{abstract}

\section{Introduction}
High-definition (HD) maps are considered pivotal elements in ensuring safe and effective navigation for autonomous vehicles \cite{int_ref2, int_ref1, int_ref3, vectornet, LaneGCN}. 
They facilitate precise planning and obstacle avoidance by providing detailed positional and semantic information about map instances.
HD map has traditionally been constructed offline utilizing Simultaneous Localization and Mapping (SLAM)-based methods \cite{Lego-loam, Lio-sam, LOAM}, involving complex processes that require significant labor intensity and economic costs. In addition, this approach is limited in its ability to provide timely updates in response to changing road conditions. Recent research has moved towards learning-based online HD map construction using onboard sensors, focusing on the generation of local maps around an autonomous vehicle. This approach eliminates costly management of HD maps, allowing immediate updates to reflect current road conditions and expansion to new locations. 

Early works viewed the map construction as a semantic segmentation challenge \cite{ bevformer, Petrv2, BEVSegFormer, LSS, sim2real, predictingoccupancy, CoBEVT, Beverse, Cross-view} based on bird's-eye view (BEV) representation obtained from various sensors. They predicted the class label for each pixel in a raster format, avoiding the complexity of generating precise vector contours.
While this method provides semantic map information, delineating map components of various classes, it falls short in capturing the precise key locations and their structural relations. Hence, its outputs are not suitably formatted for direct application to downstream tasks, such as motion forecasting \cite{r-pred, lapred} and planning \cite{nuplan}. 
To overcome this limitation, online generation of vectorized HD map have been studied in \cite{PivotNet, HdMapNet, MapTR, MapTRv2, VectorMapNet, InstaGraM, InsightMapper, MapVR}. These methods are capable of directly producing vectorized map entities, a common feature in HD maps.

\begin{figure}[!t]
    \centering
    \includegraphics[width=\textwidth]{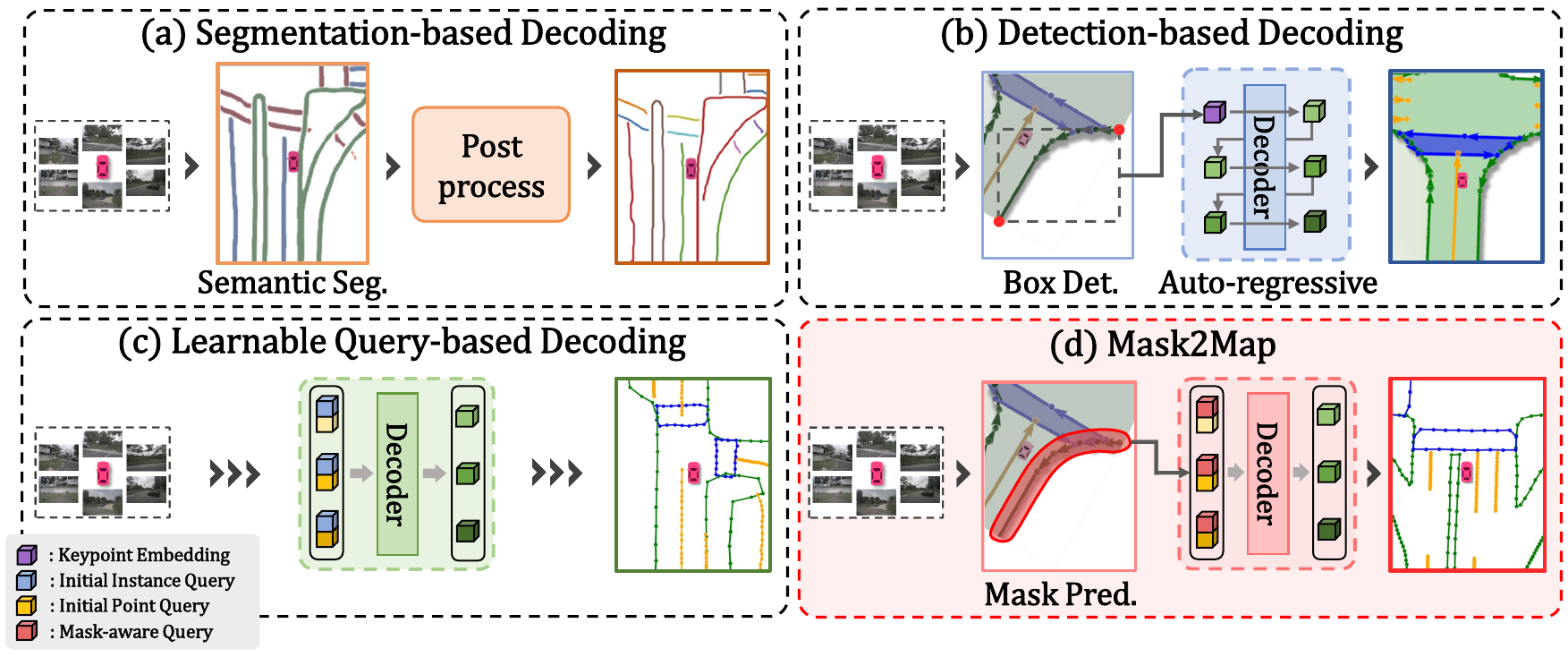}
    \caption{\textbf{Comparison of several online HD map construction methods:} (a) Segmentation-based decoding,  (b) detection-based decoding,  (c) learnable query-based decoding,  (d) proposed Mask2Map. Our Mask2Map utilizes Mask-Aware Queries to capture global-scale semantic information about a scene, enabling the generation of vectorized HD map components through subsequent query decoding.
    }
    \label{intro}
\end{figure}

To date, various online HD map construction methods have been proposed. As depicted in Figure \ref{intro} (a), the segmentation-based decoding method \cite{HdMapNet} was initially proposed, which involved semantic segmentation followed by the generation of vectorized maps using heuristic post-processing algorithms. However, this approach required significant processing time. 
The detection-based decoding method  \cite{VectorMapNet} identified key points corresponding to various instances and then generated vectorized map components sequentially, as shown in Figure \ref{intro} (b).
Nonetheless, relying solely on key points may not adequately capture the diverse shapes of instances, thus hindering the generation of accurate HD maps. Recently, various learnable query-based decoding methods were proposed \cite{PivotNet, MapTR, MapTRv2, InsightMapper, MapVR}, which directly predicted vectorized map components by decoding learnable queries from the BEV features in parallel, as illustrated in Figure \ref{intro} (c). Since initial learnable queries are unrelated to a given scene, they restrict the ability to simultaneously capture the semantic and geometric information of map instances in complex scenes.

In this study,  we introduce a novel end-to-end HD map construction framework, referred to as Mask2Map. As illustrated in Figure \ref{intro} (d), Mask2Map distinguishes itself from existing approaches by leveraging segmentation masks designed to differentiate between different classes of instances in the BEV domain.
The proposed Mask2Map architecture comprises two networks: an {\it Instance-Level Mask Prediction Network} (IMPNet) and a {\it Mask-Driven Map Prediction Network} (MMPNet). Initially, IMPNet constructs Multi-scale BEV Features from sensor data and generates Mask-Aware Queries to capture the semantic features of instances from a global perspective. Following the framework of the instance segmentation model, Mask2Former \cite{Mask2Former}, we devise Mask-Aware Queries capable of generating BEV Segmentation Masks associated with instances of different classes in the BEV domain. Subsequently, based on the Mask-Aware Queries provided by IMPNet, MMPNet dynamically predicts the ordered point set of map instances from a local perspective in the BEV domain. 
In a nutshell, MMPNet focuses on the generation of coherent and refined map components by leveraging comprehensive semantic scene information obtained from IMPNet.

We introduce several innovative approaches to enhance accuracy in predicting HD maps. First, we devise the {\it Positional Query Generator} (PQG), which generates instance-level positional queries capturing comprehensive location information to enhance Mask-Aware Queries. 
Second, while most existing methods construct the HD map without considering the point-level information of each map instance, we introduce the {\it Geometric Feature Extractor} (GFE) to capture the geometric structure for each instance. GFE processes the BEV Segmentation Masks to extract point-level geometrical features for map instances from BEV features.
Third, we observe limited performance in Mask2Map due to inter-network inconsistency when the queries from IMPNet and those from MMPNet are associated with GTs from different instances.
To address this problem, we propose an {\it Inter-network Denoising Training} strategy \cite{dn-detr, MPFormer}. This approach utilizes noisy GT queries and perturbed GT Segmentation Masks as input to IMPNet and guides the model to counteract the noise, thereby ensuring inter-network consistency and enhancing the performance of HD map construction.

We evaluate the proposed Mask2Map on multiple challenging benchmarks, including nuScenes \cite{nuscenes} and Argoverse2 \cite{argoverse2}. Our Mask2Map achieves remarkable performance gains over the previous state-of-the-art (SOTA) methods on both benchmarks. In particular, on nuScenes benchmark, Mask2Map achieves 71.6\% $mAP$, outperforming previous best camera-based method MapTRv2 \cite{MapTRv2} by 10.1\% $mAP$. In the rasterization-based evaluation metric\cite{MapVR}, Mask2Map achieves 54.7\% $mAP$ SOTA performance, more than 18.0\% $mAP$ higher than MapTRv2. On Argoverse2 benchmark, Mask2Map outperforms MapTRv2 by 4.1\% $mAP$ with the same backbone ResNet50 \cite{resnet}. 

The contributions of this study are summarized as follows:
\begin{itemize}
    \item We present Mask2Map, a new framework for online HD map construction. Our model captures semantic information at the instance-level from the scene and uses it to generate fine-grained map components subsequently. We integrate a key element from Mask2Former \cite{Mask2Former}: the Mask-Aware Query, redesigned to extract semantic masks in the BEV domain.\\
    
    \item We design a mask-guided hierarchical feature extraction architecture to efficiently encode instance-level positional information and point-level geometric information of map instances. \\

    \item We present an Inter-network Denoising Training strategy that uses noisy GT queries and perturbed GT Segmentation Masks to ensure inter-network consistency and boost the performance of HD map construction. 
\end{itemize}

\section{Related Works}
\noindent\textbf{BEV Segmentation Methods.} 
The BEV segmentation task refers to the task of gathering information about the static environment surrounding a vehicle using sensor data. Recently, many BEV segmentation methods have adopted learning-based approaches, utilizing robust deep learning backbone models developed for 3D perception \cite{ bevformer, Petrv2, BEVSegFormer, LSS, sim2real, predictingoccupancy, CoBEVT, Beverse, Cross-view}.

These methods typically extract BEV features from sensor data and perform semantic segmentation on the BEV domain using rasterized images of static scenes as GT. 
Lift-Splat-Shoot (LSS) \cite{LSS} transformed features extracted from multi-view cameras into 3D features using predicted depth information and then generated BEV representation by pooling these features.
CVT \cite{Cross-view} used cross-view attention to learn geometric transformations from perspective view to BEV domain using camera-aware positional embedding. 
BEVFormer \cite{bevformer} modeled BEV representations unified by interacting with spatial and temporal information through predefined grid-shaped BEV queries.  
BEVSegFormer \cite{BEVSegFormer} conducted BEV semantic segmentation by employing a deformable cross-attention module, which generated dense semantic queries from multi-view camera features without relying on camera intrinsic and extrinsic parameters.

\noindent\textbf{Vectorized HD Map Construction Methods.}
Recently, online HD map construction methods have received much attention for their potential to replace  HD maps in autonomous driving and provide useful information for robot planning and localization.  These methods predicted detailed map instances surrounding an ego vehicle using sensor data in real-time \cite{PivotNet, HdMapNet, MapTR, MapTRv2, VectorMapNet, BeMapNet, InstaGraM, InsightMapper, MapVR}.

HDMapNet \cite{HdMapNet} produced vectorized HD maps using a semantic segmentation model with BEV features and a post-processing method to refine the result. However, this approach demands significant computation time.
To enhance processing efficiency, query-based methods have been introduced, which leveraged Transformer attention to decode scenes and directly predict sequences of ordered points for map instances.
VectorMapNet \cite{VectorMapNet} introduced a two-stage framework that first detects bounding boxes of map instances and then sequentially predicts the points of each instance with an auto-regressive decoder.
MapTR \cite{MapTR} leveraged the architecture of DETR \cite{deformable-detr} to represent map instances as ordered point sets and encode them using hierarchical queries for a Transformer decoder.
MapTRv2 \cite{MapTRv2} further extended its capability by using depth supervision to learn 3D geometric information and conducting semantic segmentation on both perspective views and BEV.
MapVR \cite{MapVR} generated a vectorized map for each map instance and subsequently transformed it into a rasterized map using a differentiable rasterizer, providing instance-level segmentation supervision.
PivotNet \cite{PivotNet} predicted map instances by generating an ordered list of pivotal points that are crucial for capturing the overall shapes of map components.

\noindent\textbf{Denoising Training Strategy.}
Perception models based on the DETR architecture \cite{detr, MaskFormer, Mask2Former, dab-detr, conditional-detr, deformable-detr} have adopted query-based prediction using Transformer architecture, assigning GT labels to predictions via bipartite matching to ensure proper supervision. However, such assignments can occasionally result in inconsistencies in matching across epochs or layers \cite{dn-detr, MPFormer}. For instance, different GT labels may be assigned to the same query over different epochs, consequently resulting in slower convergence and decreased performance.

To address this challenge, DN-DETR \cite{dn-detr} introduced a denoising training strategy.  This strategy integrates queries derived from noisy ground truth (GT) bounding boxes into the existing queries of the DETR decoder, assigning the task of predicting GT bounding boxes from these GT queries. This approach has proven effective in stabilizing bipartite matching across training epochs. 
MP-Former \cite{MPFormer} addressed the issue of inconsistent mask predictions occurring between consecutive decoder layers. 
MP-Former employed a mask-piloted training approach that utilized both GT queries and GT masks intentionally corrupted with noise to alleviate the negative impact of inaccurate mask predictions.
Mask DINO \cite{MaskDINO} introduced a unified denoising training framework that enhanced the stability of multi-task learning for object detection and segmentation tasks.

\begin{figure}[!t]
    \centering
    \includegraphics[width=0.97\textwidth]{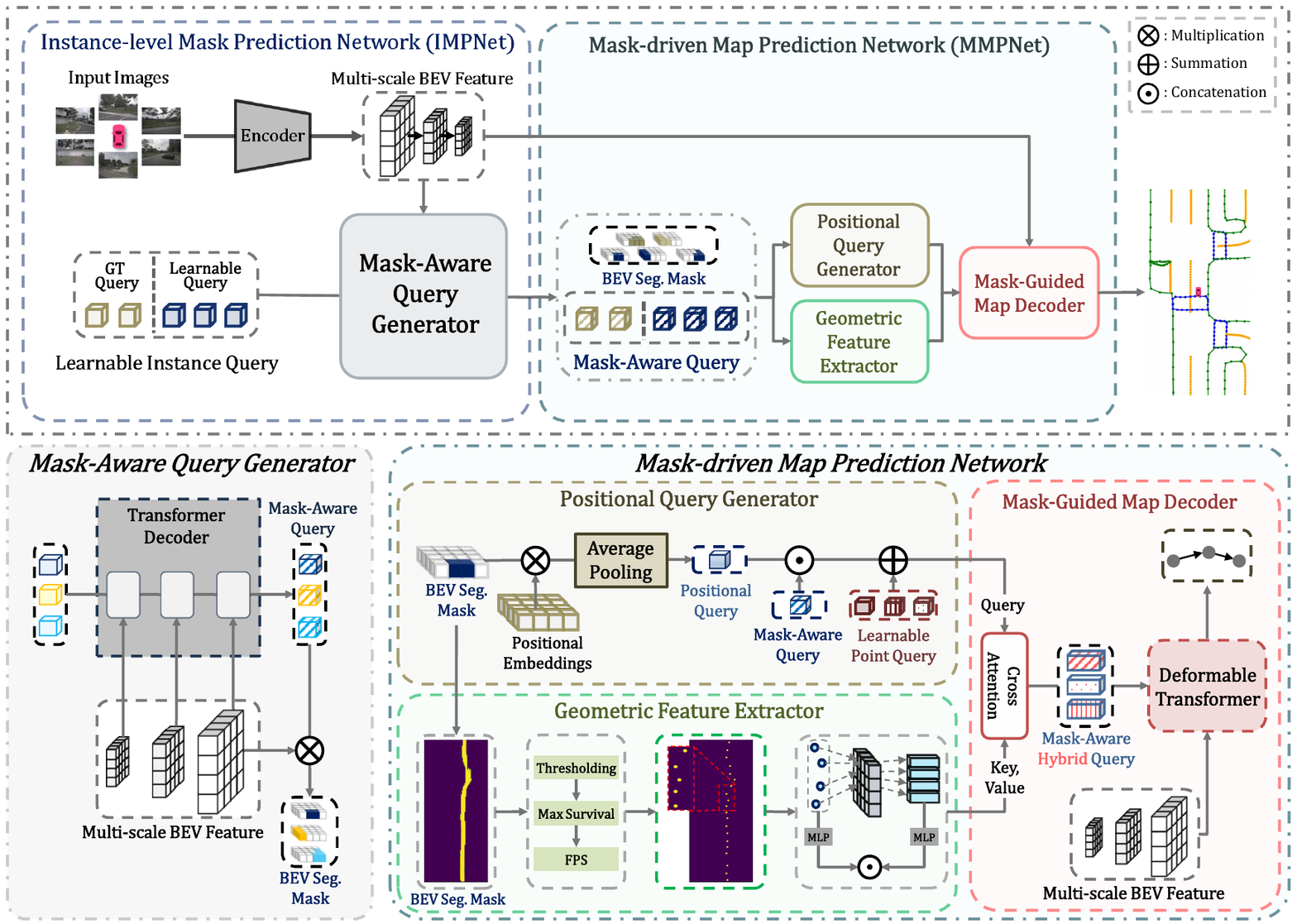}
    \caption{\textbf{Overall structure of Mask2Map.} 
    The Mask2Map system consists of IMPNet and MMPNet. IMPNet generates Mask-Aware Queries and BEV Segmentation Masks using Multi-scale BEV Features extracted from sensor data. Then, MMPNet predicts the class and ordered point set of map instances using PQG, GFE, and Mask-Guided Map Decoder. Both PQG and GFE generate semantic geometrical features on the map instances, and the Mask-Guided Map Decoder constructs vectorized maps based on these features.
    }
    \label{overall_architecture}
\end{figure}

\section{Method}

In this section, we present the details of the proposed HD map construction method, Mask2Map. 




\subsection{Overview}

An overall architecture of the Mask2Map is depicted in Figure \ref{overall_architecture}. The Mask2Map architecture comprises two networks: IMPNet and MMPNet. First, IMPNet generates Mask-Aware Queries capturing holistic semantic information from a global perspective. 
Subsequently, MMPNet constructs a more detailed vectorized map from a local perspective using geometric information acquired through PQG and GFE.


\subsection{Instance-Level Mask Prediction Network (IMPNet)}
IMPNet consists of BEV Encoder and Mask-Aware Query Generator.
BEV Encoder extracts Multi-scale BEV Features from the sensor data and Mask-Aware Query Generator produces Mask-Aware Queries, which are subsequently used to generate BEV Segmentation Masks.

\textbf{BEV Encoder.}
IMPNet generates BEV features by processing multi-view camera images, LiDAR point clouds, or a fusion of both modalities. 
Multi-view camera images are transformed into BEV representation utilizing the LSS operation \cite{LSS}. LiDAR point clouds are converted into BEV representation through voxel encoding \cite{SECOND}. When integrating both camera and LiDAR sensors for fusion, BEV features extracted from both modalities are concatenated and passed through additional convolutional layers.

Next, BEV Encoder produces BEV features of multiple scales through downsampling layers. These multi-scale features are then jointly encoded using the Deformable Transformer Encoder \cite{deformable-detr} to encode relations between Multi-scale BEV Features. 
This process yields Multi-scale BEV Features $\textbf{F}^{\text{BEV}} = \{F^{\text{BEV}}_l\}_{l=1}^S$, where  \(l\) denotes the scale index and \(S\) represents the total number of scales. 
The scale index of $l=1$ represents the smallest scale, whereas $l=S$ signifies the largest scale. We denote \(H\) and \(W\) as the height and width of the BEV feature $F^{\text{BEV}}_S$ at the largest scale.

\textbf{Mask-Aware Query Generator.}
Mask-Aware Query Generator extracts Mask-Aware Queries from Multi-scale BEV Features using the Mask Transformer proposed in Mask2Former \cite{Mask2Former}. 
The Mask-Aware Queries are initialized with learnable vectors and are decoded through $M$ layers of the Transformer decoder. 
Given Multi-scale BEV Features $\mathbf{F}^{\text{BEV}}$ and the BEV Segmentation Masks $\mathbf{M}_{m-1}=\{M_{i,{m-1}}\}_{i=1}^{N_I}$ obtained at the $(m-1)$-th decoding layer,  the Mask-Aware Queries $\mathbf{q}^{\text{mask}}_{m-1}=\{q^{\text{mask}}_{i,{m-1}}\}_{i=1}^{N_I}$  are updated as 
 \begin{gather}
\hat{\mathbf{M}}_{m-1} = 
\begin{cases} 
0 & \text{if }\mathbf{M}_{m-1} > \tau_M \\
-\infty & \text{otherwise}
\end{cases} \\
Q_{m} = \mathbf{q}^{\text{mask}}_{m-1} W^{Q},\quad K_{m} = F^{\text{BEV}}_m W^{K},\quad V_{m} = F^{\text{BEV}}_m W^{V} \\[8pt]
  \mathbf{q}^{\text{mask}}_{m} = \text{softmax}(\hat{\mathbf{M}}_{m-1} + Q_{m}K_{m}^T)V_{m}+{\mathbf{q}}^{\text{mask}}_{m-1}, 
\end{gather}
where \( {\tau}_M \) denotes a threshold, $N_I$ denotes the number of the Mask-Aware Queries, and $W^{Q}$, $W^{V}$, and $W^{K}$ are learnable weight matrices. 
Finally, the BEV Segmentation Masks $\mathbf{M}_{m}$ are obtained by applying dot product between BEV feature $F^{\text{BEV}}_S$ of the largest scale and the Mask-Aware Queries $\mathbf{q}^{\text{mask}}_{m}$ along the channel axis. 
Then the sigmoid function is applied to normalize the BEV Segmentation Masks. These BEV Segmentation Masks are then fed back into the next decoding layer for further refinement.  
After $M$ decoding layers, IMPNet ends up with the final Mask-Aware Queries $\mathbf{q}^{\text{mask}}=\mathbf{q}^{\text{mask}}_{M}$ and the BEV Segmentation Masks $\textbf{M}^{\text{BEV}}=\mathbf{M}_{M}$, which are delivered to the subsequent MMPNet. 

\subsection{Mask-Driven Map Prediction Network (MMPNet)}
MMPNet comprises three main components: the Positional Query Generator, the Geometric Feature Extractor, and the Mask-Guided Map Decoder.
The Positional Query Generator injects positional information to enhance Mask-Aware Queries, while the Geometric Feature Extractor processes the BEV Segmentation Masks to extract geometrical features from the BEV feature. Finally, the Mask-Guided Map Decoder predicts both the class and coordinates of ordered points for map instances using the features provided by the Positional Query Generator and Geometric Feature Extractor.

\textbf{Positional Query Generator.} 
While Mask-Aware Queries carry semantic information about map instances, they lack positional information. To enable MMPNet to generate coordinates of points for map instances, it is essential to integrate positional information in the BEV domain into the Mask-Aware Queries.
PQG initially derives the sparsified BEV mask from the BEV Segmentation Mask $\textbf{M}^{\text{BEV}}$,
\begin{gather}
\hat{\mathbf{M}}^{\text{PQG}} = 
\begin{cases} 
\mathbf{M}^{\text{BEV}} & \text{if } \mathbf{M}^{\text{BEV}} > {\tau}_P \\
0 & \text{otherwise}
\end{cases} .
\end{gather}
PQG injects the 2D positional embedding ${PE}$ to the sparsified BEV mask $\hat{\mathbf{M}}_i^{\text{PQG}}$, where $PE$ is generated by sinusoidal functions  \cite{detr}.  Then, the positional queries $f^{\text{pos}}_i$ is obtained by applying average pooling in both $x$ and $y$ domains, i.e., 
\begin{gather}
 f^{\text{pos}}_i = \frac{1}{N_{\text{nz}}^i}\sum_{x=1}^{H}\sum_{y=1}^{W} {(\hat{\mathbf{M}}_i^{\rm PQG}(x,y) \otimes {PE}(x,y))}, 
\end{gather}
where $i\in [1,N_I]$,  \(N_{\text{nz}}^i\) denotes the number of non-zero pixels in \(\hat{\mathbf{M}}_i^{\rm PQG}\) and \(\otimes\) is the element-wise product. 
The positional queries $\textbf{f}^{\text{pos}}$ are concatenated with the Mask-Aware Queries $\textbf{q}^{\text{mask}}$ to generate the Combined Positional Queries $\hat{\textbf{f}}^{\text{pos}}$.
Next, the Combined Positional Query {$\hat{f}_i^{\text{pos}}$ is used to produce $N_P$ point-level features for the $i$-th map instance.
Towards this goal, PQG replicates $\hat{f}_i^{\text{pos}}$ $N_P$ times and adds them with $N_P$ Learnable Point Queries $q_1$, ..., $q_{N_P}$, 
generating the PQG Query Features $\mathbf{q}^{\text{PQG}}_{i} = \{q^{\text{PQG}}_{i,j}\}_{j=1}^{N_P}$,
\begin{gather}
q^{\text{PQG}}_{i,j} = \hat{f}_i^{\text{pos}} + q_j,
\end{gather}
where $i\in [1,N_I]$ and $j\in [1,N_P]$. 
Note that the learnable queries $q_1$ through $q_{N_P}$ give the Mask-Aware Queries an idea of the sequential order of points generated for the $i$-th map instance.
The resulting  PQG Query Features $\mathbf{q}_{i}^{\text{PQG}}$ are delivered to the Mask-Guided Map Decoder. 

\textbf{Geometric Feature Extractor.}
GFE generates point-wise features that capture the geometric structure of map instances. Initially, using the threshold $\tau_G$, GFE produces the sparsified BEV mask $\hat{\textbf{M}}^{\text{GFE}}$ from the BEV Segmentation Mask $\textbf{M}^{\text{BEV}}$.
To generate point-wise geometric features, GFE samples $N_S$ key pixels from the sparsified BEV mask $\hat{M}_i^{\text{GFE}}$.
First, we employ the {\it Max Survival} method, which selects the strongest pixel from the non-overlapping window of size $G \times G$ sliding on $\hat{M}_i^{\text{GFE}}$ while setting the remaining pixels to zero.
Next, we apply the Farthest Point Sampling (FPS) method \cite{pointnet++} to iteratively select the output of the Max Survival method and identify $N_S$ key points.
Finally, based on the positions of the $N_S$ key points, $N_S$ features are pooled from the BEV features $F^{\text{BEV}}_S$ of the largest scale. Concurrently, the $(x,y)$ coordinates of these $N_S$ key points are encoded using a Multi-Layer Perceptron (MLP). These two features are concatenated, resulting in the GFE Features denoted as $\mathbf{f}_{i}^{\text{GFE}} = \{f_{i, j}^{\text{GFE}}\}_{j=1}^{N_S}$.

\textbf{Mask-Guided Map Decoder.}
The Mask-Guided Map Decoder predicts the class and sequence of ordered points for vectorized map components based on PQG Query Features $\mathbf{q}_{i}^{\text{PQG}}$ and GFE Features $\textbf{f}_{i}^{\text{GFE}}$. By using $\mathbf{q}_{i}^{\text{PQG}}$ as queries and $\textbf{f}_{i}^{\text{GFE}}$ as keys and values, the cross-attention module produces Mask-Aware Hybrid Queries $\mathbf{q}_{i}^{\text{Hybrid}} = \{q_{i, j}^{\text{Hybrid}}\}_{j=1}^{N_P}$. These queries are subsequently decoded by a Deformable Transformer \cite{deformable-detr}, utilizing the Multi-scale BEV Features $\textbf{F}^{\text{BEV}}$ as values. 
Finally, the prediction heads predict instance classification scores and normalized BEV coordinates for each map instance through the classification and regression heads, respectively.

\begin{figure}[!t]
    \centering
    \includegraphics[width=0.85\textwidth]{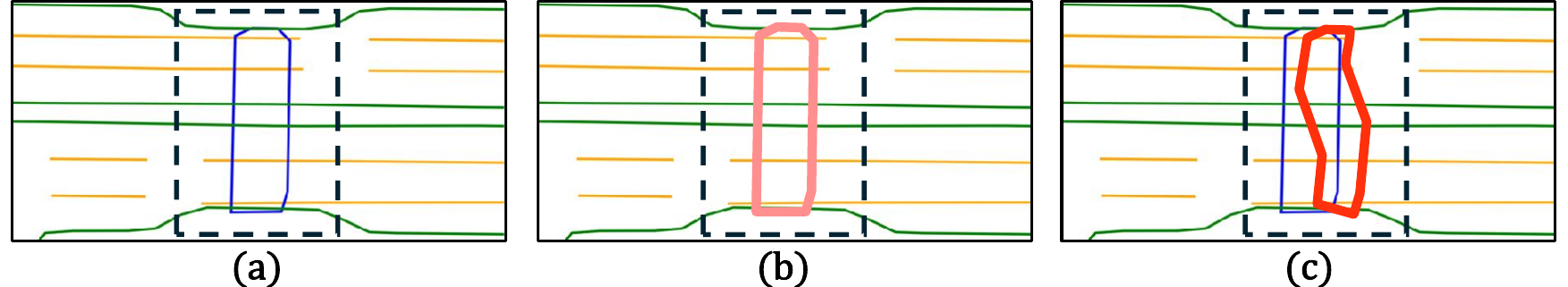}
    \caption{\textbf{Illustration of proposed Map Noise method.} (a) The blue polygon denotes a vectorized GT of a pedestrian crossing. (b) The pink polygon represents a GT Segmentation Mask without noise. (c) The red polygon represents the result of adding Map Noise to the GT Segmentation Mask.
    }    
    \label{noise_plot}
\end{figure}

\subsection{Inter-network Denoising Training}

Mask2Map passes Mask-Aware Queries from IMPNet to MMPNet for hierarchical refinement of instance features. To ensure efficient training, we assign instance segmentation loss for IMPNet and map construction loss for MMPNet. Following the training strategy suggested in \cite{detr}, queries used by IMPNet and those by MMPNet should be matched to their respective GT through bipartite matching. However, inconsistencies in this matching process can occur when the queries used in both IMPNet and MMPNet are matched with GTs associated with different instances. 
We observe that this inter-network inconsistency tends to cause slower convergence and diminished performance.

To address this issue, we adopt a denoising training strategy \cite{MPFormer}. 
The key idea is to merge noisy GT Queries, which are derived from each GT instance, into the learnable queries within IMPNet. (see Figure \ref{overall_architecture}). Our model is trained to denoise these queries by directly matching them with their corresponding GTs.
 This is contrasted with the learnable queries, which are matched to the GTs through bipartite matching. Thus, this strategy is called Inter-network Denoising Training.  This process guides the model to establish a correspondence between the queries used in IMPNet and MMPNet, effectively mitigating inter-network inconsistency. Additionally, alongside GT Queries, we also generate GT Segmentation Masks, which replace the BEV Segmentation Masks for IMPNet.

We generate GT Queries by assigning one of the $C$ class embedding vectors corresponding to a GT class of each instance, where $C$ denotes the number of classes. 
We add a flipping noise by randomly replacing the class embedding vector with one of other classes at a probability of $\lambda$.
Simultaneously, we also generate perturbed GT Segmentation Masks by adding 
{\it Map Noises} to a sequence of ordered points of each instance and rasterize them, as shown in Figure \ref{noise_plot} (c). 
 
The combination of the noisy GT Queries and learnable queries is referred to as Learnable Instance Queries.  
Instead of utilizing BEV Segmentation Masks, we exclusively employ the perturbed GT Segmentation Masks for noisy GT Queries. 
The noisy GT Queries pass through both IMPNet and MMPNet and their prediction results are matched with the corresponding GTs without the bipartite matching. 

\subsection{Training Loss}
The total loss $L$ used to train Mask2Map is given by 
\begin{align}
\mathcal{L} = \mathcal{L}_{\text{seg}} + \mathcal{L}_{\text{map}} + \mathcal{L}_{\text{aux}} + \mathcal{L}_{\text{dn}},
\end{align}
where $\mathcal{L}_{\text{seg}}$ is the loss term for training IMPNet on a BEV segmentation task, $\mathcal{L}_{\text{map}}$ is the loss term for training MMPNet on a map construction task, $\mathcal{L}_{\text{aux}}$ is the auxiliary loss term, $\mathcal{L}_{\text{dn}}$ is the loss term for Inter-network Denoising Training.

We use bipartite matching with Hungarian solver to assign the queries used by IMPNet and those by MMPNet to their respective GTs. Based on the assignment, we calculate the $\mathcal{L}_{\text{seg}}$ and $\mathcal{L}_{\text{map}}$.
We adopt \cite{Mask2Former} to obtain the loss term $\mathcal{L}_{\text{seg}}$. 
The loss term $\mathcal{L}_{\text{map}}$ consists of L1 loss for regression of vectorized map instances, focal loss\cite{focal} for the classification of instances, and cosine similarity loss between the direction calculated from adjacent points of GT and the one from the predictions. 
The auxiliary loss term $\mathcal{L}_{\text{aux}}$ calculates the error for depth estimation and 2D map semantic segmentation tasks conducted on the camera perspective-view features  \cite{MapTRv2}. 
The loss term \(\mathcal{L}_{\text{dn}}\) is the summation of two terms, each corresponding to the loss used in IMPNet and MMPNet without bipartite matching between GT and predictions for noisy GT Queries.

\begin{table*} [t!] 
    \caption{Comparison with SOTA methods on the nuScenes validation set. FPSs are measured on the same machine equipped with RTX3090. The "-" denotes that the associated results are not available. "C" and "L" respectively denote camera and LiDAR. The "R50", "PP" and "Sec" respectively correspond to ResNet50\cite{resnet}, PointPillars\cite{pointpillars}, and SECOND\cite{SECOND}.} 
    \centering
    \begin{adjustbox}{width=0.8\textwidth}
    \begin{tabular}{l|ccc|cccc|cc}
    \toprule[1.2pt]
    Method & Modality & Backbone & Epoch & $AP_{ped}$ & $AP_{divider}$ & $AP_{boundary}$ & $mAP$ & FPS  \\ \toprule[0.7pt]
    {MapTR\cite{MapTR}} & C & R50 & 24 & 46.3 & 51.5 & 53.1 & 50.3 & 15.1  \\
    {MapVR\cite{MapVR}} & C & R50 & 24 & 47.7 & 54.4 & 51.4 & 51.2 & 15.1 \\
    {PivotNet\cite{PivotNet}} & C & R50 & 24 & 56.2 & 56.5 & 60.1 & 57.6 & - \\ 
    {BeMapNet\cite{BeMapNet}}  & C & R50 & 30 & 57.7 & 62.3 & 59.4 & 59.8 & - \\ 
    {MapTRv2\cite{MapTRv2}}& C  & R50 & 24 & 59.8 & 62.4 & 62.4 & 61.5 & 14.1 \\ 
    \hline \hline 
    {Ours} & C & R50 & 24 & \textbf{70.6} & \textbf{71.3} & \textbf{72.9} & \textbf{71.6} & 10.1 \\ \bottomrule[0.7pt]
     {VectorMapNet\cite{VectorMapNet}} & C & R50 & 110 & 36.1 & 47.3 & 39.3 & 40.9 & 2.2  \\ 
     {MapTR\cite{MapTR}} & C & R50 & 110 & 56.2 & 59.8 & 60.1 & 58.7 & 15.1  \\
     {MapVR\cite{MapVR}} & C & R50 & 110 & 55.0 & 61.8 & 59.4 & 58.8 & 15.1 \\
     {MapTRv2\cite{MapTRv2}} & C & R50 & 110 & 68.1 & 68.3 & 69.7 & 68.7 & 14.1 \\ \bottomrule[0.5pt]
    {Ours} & C & R50 & 110 & \textbf{73.6} & \textbf{73.1} & \textbf{77.3} & \textbf{74.6} & 10.1 \\ \bottomrule[0.7pt]
     {VectorMapNet\cite{VectorMapNet}} & C+L & R50 \& PP & 110 & 37.6 & 50.5 & 47.5 & 45.2 & - \\ 
     {MapTR\cite{MapTR}} & C+L & R50 \& Sec & 24 & 55.9 & 62.3 & 69.3 & 62.5 & 6.0 \\ 
     {MapVR\cite{MapVR}} & C+L & R50 \& Sec & 24 & 60.4 & 62.7 & 67.2 & 63.5 & 6.0 \\
     {MapTRv2\cite{MapTRv2}} & C+L & R50 \& Sec & 24 & 65.6 & 66.5 & 74.8 & 69.0 & 5.8 \\\hline \hline
    {Ours} & C+L & R50 \& Sec & 24 & \textbf{76.5} & \textbf{76.6} & \textbf{82.1} & \textbf{78.4} & 4.1 \\ 
    \bottomrule[1.2pt]
    \end{tabular}
    \end{adjustbox}
    \label{nusc_quantitative}
\end{table*}

\begin{table}[!t]
\begin{minipage}[t]{0.49\linewidth}
  \centering
      \caption{Comparison of SOTA methods on nuScenes validation set with rasterization-based metric. All models use camera modality for input and ResNet50 \cite{resnet} as backbones. The "$*$" indicates results reproduced using public codes. }
    \centering
    \begin{adjustbox}{width=1.0\textwidth}
    \begin{tabular}{l|c|cccc}
    \toprule[1.2pt]
    Method & Epoch & $AP_{ped}^{\dagger}$ & $AP_{divider}^{\dagger}$ & $AP_{boundary}^{\dagger}$ & $mAP^{\dagger}$  \\ \toprule[0.7pt]
    MapTR\cite{MapTR} &24& 32.4 & 23.5 & 17.1 & 24.3  \\
    MapVR\cite{MapVR} &24&37.5 & 33.1 & 23.0 & 31.2 \\
    $\text{MapTRv2}^{*}$\cite{MapTRv2} &24& 49.9 & 34.7 & 25.7& 36.7  \\ \hline \hline 
    Ours &24& \textbf{62.9} & \textbf{52.3} & \textbf{48.9} & \textbf{54.7} \\ \bottomrule[1.2pt]
    \end{tabular}
    \end{adjustbox}
    \label{nusc_quantitative_rasterized}


\end{minipage}%
\hspace{1mm}
\begin{minipage}[t]{0.49\linewidth}
  \centering
      \caption{Comparison with state-of-the-art methods on the Argoverse2 validation set. All of the presented results derive from models trained using camera data as the input. The "EB0" correspond to EfficientNet-B0 \cite{efficientnet}} 
    \centering
    \begin{adjustbox}{width=\textwidth}
    \begin{tabular}{l|c|cccc}
    \toprule[1.2pt]
    Method  & Backbone & $AP_{ped}$ & $AP_{divider}$ & $AP_{boundary}$ & $mAP$  \\ \toprule[0.7pt]
    HDMapNet\cite{HdMapNet} & EB0 & 13.1 & 5.7 & 37.6 & 18.8 \\
    VectorMapNet\cite{VectorMapNet} &R50 & 38.3 & 36.1 & 39.2 & 37.9  \\
    MapTR\cite{MapTR}&R50  & 54.7 & 58.1 & 56.7 & 56.5  \\
    MapVR\cite{MapVR}&R50  & 54.6 & 60.0 & 58.0 & 57.5  \\
    MapTRv2\cite{MapTRv2}&R50  & 62.9 & 72.1 & 67.1 & 67.4  \\ \hline \hline
    Ours&R50  & \textbf{68.1} & \textbf{72.7} & \textbf{73.7} & \textbf{71.5}  \\ \bottomrule[1.2pt]
    \end{tabular}
    \end{adjustbox}
    \label{av2_quantitative}

\end{minipage}
\end{table}

\section{Experiments}
\subsection{Experimental Settings}
\noindent\textbf{Datasets.}
Both nuScenes \cite{nuscenes} and Argoverse2 \cite{argoverse2} datasets are real-world driving datasets that provide HD map labels annotated by hand. 
The nuScenes dataset consists of $28K$ frames for training set and $6K$ frames for validation set. 
Each frame is annotated at $2$Hz from six surrounding RGB cameras and a 32-beam LiDAR covering 360$^{\circ}$ field of view.
The Argoverse2 dataset contains training set of $110K$ frames and validation set of $24K$ frames. Annotations for keyframes are presented at 10hz with 7 ring cameras and two 32-beam LiDARs.

\noindent\textbf{Evaluation Metrics.}
We define the perception range $[-15.0m, 15.0m]$ for the lateral direction and $[-30.0m, 30.0m]$ for the longitudinal direction. Following prior works \cite{PivotNet,MapTR,MapTRv2,VectorMapNet,MapVR, InsightMapper}, we categorize map instances into three types for HD map construction: {\it Pedestrian Crossing}, {\it Lane Divider}, and {\it Road Boundary}. 
We adopt two evaluation metrics: Average Precision (AP) based on Chamfer distance proposed in \cite{HdMapNet} and AP based on rasterization proposed in \cite{MapVR}.
We primarily utilize the Chamfer distance metric, using thresholds of $0.5$, $1.0$, and $1.5$ meters for mean AP ($mAP$).
For rasterization-based mean AP ($mAP^{\dagger}$), we measure intersection over union for each map instance, with thresholds set $\{0.50, 0.55, ..., 0.75\}$ for pedestrian crossings and $\{0.25, 0.30, ..., 0.50\}$ for line-shaped elements.
To further evaluate the matching consistency ratio of inter-network, we use the Query Utilization (\(Util\)) metric proposed by \cite{MPFormer}, which calculates the consistency ratio of GT matches in MMPNet's first decoder layer with the matches in IMPNet's last layer.

\noindent\textbf{Implementation Details.}
We adopted ResNet50 \cite{resnet} for image backbone networks. For nuScenes, images with dimensions of 1600$\times$900 were resized by a $0.5$ ratio. In the case of Argoverse2, seven images with dimensions of 1550$\times$2048 for the front view, and others with dimensions of 2048$\times$1550 were padded to 2048$\times$2048 before being resized by a $0.3$ ratio.
The LiDAR point clouds were voxelized with a voxel size of $0.1$, $0.1$, and $0.2$. The voxel features were extracted by SECOND \cite{SECOND}.
We employed six BEV encoder layers and three mask Transformer layers in IMPNet. We employed six Transformer decoder layers in MMPNet.
The thresholds for BEV Segmentation Mask, $\tau_M$, $\tau_P$, and $\tau_G$ were set to $0.5$, $0.3$, and $0.8$, respectively. 
We configured the number of instance queries to 50 and the number of point queries to 20.
In GFE, we set the window size ($G$) for the Max Survival method to $4$ and the number of sampling points ($N_S$) to $20$. 
The flipping noise probability $\lambda$ was set to $0.2$.
For optimization, we employed AdamW with a weight decay of 0.01 and utilized cosine annealing as a scheduler. The initial learning rate was set to 6e-4.  Our model was trained on 4 RTX3090 GPUs with batch size $4$ per GPU.

\subsection{Performance Comparison}
\noindent\textbf{Results on nuScenes.}
Table \ref{nusc_quantitative} presents the comprehensive performance analysis of Mask2Map on the nuScenes validation set \cite{nuscenes}, utilizing the Chamfer distance metric. Mask2Map establishes a new state-of-the-art performance, exhibiting substantial improvements over existing methods \cite{PivotNet,HdMapNet,MapTR,MapTRv2,VectorMapNet,BeMapNet,MapVR}.
When using camera input only, Mask2Map achieves noteworthy results of 71.6\% $mAP$ at 24 epochs and 74.6\% $mAP$ at 110 epochs, outperforming the previous state-of-the-art model, MapTRv2 \cite{MapTRv2}, by 10.1\% $mAP$ and 5.9\% $mAP$, respectively. When using camera-LiDAR fusion, Mask2Map achieves the performance gain of 9.4\% $mAP$ over MapTRv2.  
Table \ref{nusc_quantitative_rasterized} evaluates the performance of Mask2Map based on a rasterization-based metric. Notably, our Mask2Map method achieves a remarkable performance gain of 18.0 $mAP$ over MapTRv2.

\noindent\textbf{Results on Argoverse2.}
 Table \ref{av2_quantitative} presents the performance evaluation of several HD map construction methods \cite{MapTR, MapTRv2, VectorMapNet, MapVR} on the Argoverse2 validation set \cite{argoverse2}.
 The proposed Mask2Map demonstrates significant performance improvements compared to existing models. Mask2Map surpasses the current leading method, MapTRv2, by 4.1\% $mAP$, demonstrating that our model achieves the consistent performance across different scenarios.

\begin{table}[t!]
\begin{minipage}[t]{0.49\linewidth}
  \centering
  \centering
\caption{Ablation study of main components of Mask2Map} 
\begin{adjustbox}{width=0.90\textwidth}
    \begin{tabular}{l|cccc}
        \toprule[1.2pt]
         & $AP_{ped}$ & $AP_{divider}$ & $AP_{boundary}$ & $mAP$  \\ \toprule[0.7pt]
        Baseline & 30.1 & 41.5 & 46.6 & 39.4 \\
        +IMPNet& 42.7 & 45.3 & 48.1 & 45.3 \\
        +MMPNet & 45.7 & 47.0 & 54.4 & 49.1 \\
        +Denoising & \textbf{52.9} & \textbf{55.5} & \textbf{58.5} & \textbf{55.6} \\
        \bottomrule[1.2pt]
    \end{tabular}
\end{adjustbox}
\label{ablation_main}

\end{minipage}%
\hspace{1mm}
\begin{minipage}[t]{0.49\linewidth}
  \centering
  \centering
\caption{Ablation study for evaluating the contributions of MMPNet's submodules} 
\begin{adjustbox}{width=0.90\textwidth}
    \begin{tabular}{cc|cccc}
    \toprule[1.2pt]
         PQG & GFE & $AP_{ped}$ & $AP_{divider}$ & $AP_{boundary}$ & $mAP$  \\ \toprule[0.7pt]
            & & 47.6 & 51.1 & 53.8 & 50.8 \\
          & \checkmark & 49.7 & 54.2 & 57.8 & 53.9 \\
          \checkmark &  & 49.9 & \textbf{55.6} & 57.2 & 54.2 \\
         \checkmark & \checkmark  & \textbf{52.9} & 55.5 & \textbf{58.5} & \textbf{55.6} \\ \bottomrule[1.2pt]
    \end{tabular}
\end{adjustbox}
\label{ablation_module}
\end{minipage}
\end{table}

\begin{table}[t!]
\begin{minipage}[t]{0.49\linewidth}
  \centering
  \centering
\caption{Ablation study for evaluating the effect of Inter-network Denoising Training} 
\begin{adjustbox}{width=0.95\textwidth}
    \begin{tabular}{c|c|cccc}
    \toprule[1.2pt]
          Denoising & $Util$ & $AP_{ped}$ & $AP_{divider}$ & $AP_{boundary}$ & $mAP$  \\ \toprule[0.7pt]
           & 24.7 & 45.7 & 47.0 & 54.4 & 49.1 \\ 
         \checkmark & \textbf{74.7} & \textbf{52.9} & \textbf{55.5} & \textbf{58.5} & \textbf{55.6} \\ \bottomrule[1.2pt]
    \end{tabular}
\end{adjustbox}
\label{ablation_dn_matching}

\end{minipage}
\hspace{1mm}
\begin{minipage}[t]{0.49\linewidth}
  \centering
  \centering
\caption{Ablation study for Map Noise applied to GT Segmentation Masks}
\begin{adjustbox}{width=0.84\textwidth}
    \begin{tabular}{c|cccc}
    \toprule[1.2pt]
     Map Noise & $AP_{ped}$ & $AP_{divider}$ & $AP_{boundary}$ & $mAP$  \\ \toprule[0.7pt]
      & 49.7 & \textbf{56.1} & \textbf{58.6} & 54.8 \\
     \checkmark & \textbf{52.9} & 55.5 & 58.5 & \textbf{55.6} \\ \bottomrule[1.2pt]
    \end{tabular}
\end{adjustbox}
\label{ablation_noisetype}
\end{minipage}%
\end{table}

\subsection{Ablation Study}}
We conducted an ablation study to evaluate the contributions of the core ideas of Mask2Map. Camera-only input and ResNet50 \cite{resnet} backbone were used in these experiments. Training was conducted on 1/4 of the nuScenes training dataset for 24 epochs. Evaluation was performed on the entire validation set.

\noindent\textbf{Contributions of Main Components.}
Table \ref{ablation_main} demonstrates the impact of each component of Mask2Map. We evaluated performance by adding each component one by one. The first row represents a baseline model using an LSS-based BEV encoder for extracting BEV features and Deformable attention for predicting vectorized map instances \cite{MapTRv2}. 
When adding IMPNet into the baseline model, we notice a substantial 5.9\% increase in $mAP$, indicating that the inclusion of Mask-Aware Queries, capable of generating instance segmentation results, significantly boosts the performance of HD map construction.
Furthermore, the addition of MMPNet results in a further improvement of 3.8\% $mAP$, underscoring the significant contribution of injecting positional and geometric information of map instances through BEV Segmentation Masks.
Lastly, our Inter-network Denoising Training offers a 6.5\% additional increase in $mAP$, emphasizing its effectiveness in enhancing performance.

\noindent\textbf{Contributions of MMPNet's Submodules.}
In our investigation detailed in Table \ref{ablation_module}, we explored the contributions of PQG and GFE. GFE alone contributes to a notable 3.1\% increase in $mAP$ over the baseline, while PQG alone yields a 3.4\% improvement in $mAP$. The combination of PQG and GFE further enhances performance by 4.8\% $mAP$, demonstrating their complementary effects.

\begin{figure}[!t]
    \centering
    \includegraphics[width=\textwidth]{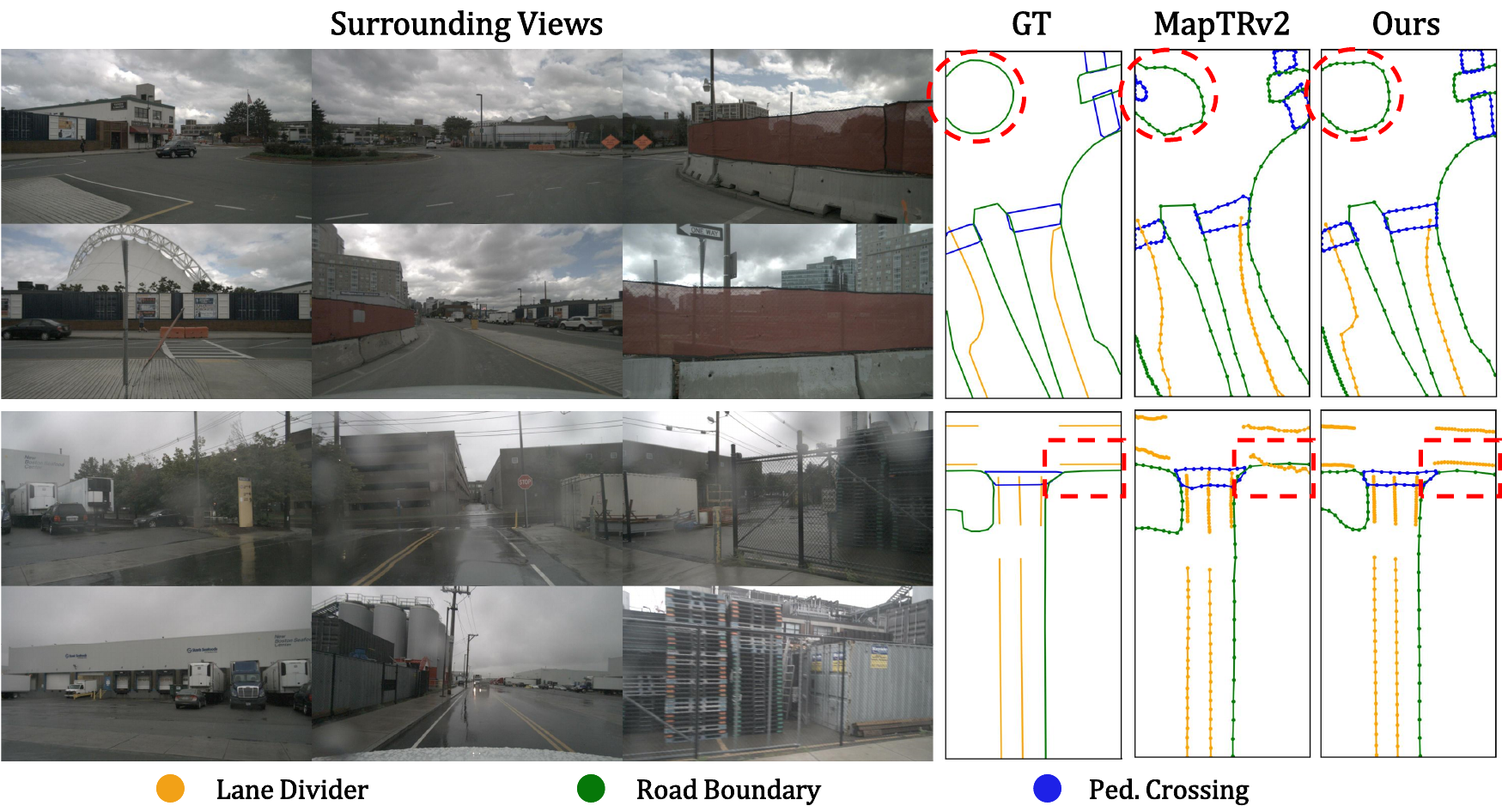}
    \caption{\textbf{Qualitative results on the nuScenes validation set.} We compared our method with  MapTRv2. The regions marked by a red ellipse and rectangle emphasize the superior results generated by our proposed model.
    }    
    \label{qualitative_plot}
\end{figure}

\noindent\textbf{Effect of Inter-network Denoising Training on Matching Consistency.}
We further investigate the impact of Inter-network Denoising Training. As depicted in Table \ref{ablation_dn_matching}, Inter-network Denoising Training drastically increases the matching ratio $Util$ from 24.7\% to 74.7\%, which is translated into a substantial 6.5\% increase in overall $mAP$ performance. This demonstrates that our Inter-network Denoising Training effectively mitigates the inconsistency in query-to-GT matching between IMPNet and MMPNet, as intended.

\noindent\textbf{Impact of Noise in Inter-network Denoising Training.}
In Table \ref{ablation_noisetype}, we explore the effect of Map Noise used in the Inter-network Denoising Training. We compare our method with a baseline using GT Segmentation Masks without Map Noise. Our findings show that adding Map Noise to GT results in a 0.8\% improvement in $mAP$ over the baseline, indicating the benefit of this approach.

\subsection{Qualitative Analysis}
\noindent\textbf{Qualitative Results.}
Figure \ref{qualitative_plot} presents the qualitative results produced by the proposed Mask2Map. We compare our method with the current SOTA, MapTRv2. 
Note that Mask2Map yields notably better map construction results than MapTRv2.
More qualitative results are provided in Supplementary Material.

\section{Discussion and Conclusion}

\noindent\textbf{Limitations and Future Work.}
As for future work, we consider improving Mask2Map in two aspects. 
(i) Temporal information is known to improve the reliability of results in autonomous driving perception tasks. However, our model currently relies solely on input from the current frame, which may lead to performance degradation in scenes occluded by objects. 
Temporal fusion methods through queries or BEV features of previous frames may provide promising paths toward addressing this limitation.
(ii) Our experiments showed that Mask2Map's FPS decreased compared to the current SOTA, MapTRv2 \cite{MapTRv2}, in exchange for substantial performance gains. To meet real-time requirements, we consider employing model compression and optimization methods. 
These techniques will be a promising avenue to improve the FPS without sacrificing performance.

\noindent\textbf{Conclusion.}
In this paper, we introduced an end-to-end online HD map construction method called Mask2Map. Mask2Map utilizes IMPNet to generate both Mask-Aware Queries and BEV Segmentation Masks, capturing semantic scene context from a global perspective. Subsequently, MMPNet enhances Mask-Aware Queries by incorporating semantic and geometrical information through PQG and GFE.
Finally, Mask-Guided Map Decoder predicts the class and ordered point set of map instances.
Additionally, we proposed Inter-network Denoising Training to mitigate inter-network inconsistency arising from differing bipartite matching results between IMPNet and MMPNet.
Our evaluation on nuScenes and Argoverse2 benchmarks demonstrated that the proposed ideas yielded significant performance improvements over the baseline, surpassing existing HD map construction methods by considerable margins.

\bibliography{mask2map}

\newpage

\appendix
\begin{center}
{\Large \textbf{Supplementary Materials for Mask2Map}}
\end{center}

\setcounter{figure}{4}
\setcounter{table}{7}
\setcounter{equation}{6}

\noindent
In this Supplementary Materials, we provide more details that could not be included in the main paper due to space limitations. We present additional experiment analysis and visualization details conducted on nuScenes and Argoverse2 benchmark datasets about the proposed Mask2Map. We discuss the following:


\begin{itemize}
    \item Additional experiment results for map construction;
    \item Extensive qualitative results of Mask2Map.
\end{itemize}

\section{Additional Experimental Results}
In this section, we provide the efficacy of Mask2Map in 3D HD map construction and present the results of additional ablation studies.

\begin{table}[h]
  \centering
\begin{minipage}[t]{0.90\linewidth}
      \caption{Comparison with the state-of-the-art methods on the Argoverse2 validation set for 3D vectorized map. All models are trained with camera modality as input.} 
    \centering
    \begin{adjustbox}{width=0.8\textwidth}
    \begin{tabular}{l|c|cccc}
    \toprule[1.2pt]
    Method  & Backbone & $AP_{ped}$ & $AP_{divider}$ & $AP_{boundary}$ & $mAP$  \\ \toprule[0.7pt]
    VectorMapNet \cite{VectorMapNet} & R50 & 36.5 & 35.0 & 36.2 & 35.8  \\
    MapTRv2 \cite{MapTRv2} &R50  & 60.7 & 68.9 & 64.5 & 64.7  \\ \hline \hline
    Ours&R50  & \textbf{67.7} & \textbf{73.0} & \textbf{73.5} & \textbf{71.4}  \\ \bottomrule[1.2pt]
    \end{tabular}
    \end{adjustbox}
    \label{av2_3d_quantitative}

\end{minipage}
\end{table}

\subsection{Comparison of 3D HD Map Construction on Argoverse2}

Table \ref{av2_3d_quantitative} presents the performance evaluation of 3D HD map construction methods on the Argoverse2 validation set.
These methods generate map instances as sets of ordered 3D points, with each point represented by its \((x, y, z)\) coordinates.
Our model achieves a new state-of-the-art performance of 71.4\% $mAP$ in 3D HD map construction on the Argoverse2 validation set, which is 6.7\% $mAP$ higher than MapTRv2 \cite{MapTRv2}.
This result demonstrates the advanced generalization capabilities of Mask2Map.

\subsection{Additional Ablation Study on nuScenes}
We conducted additional ablation studies for a detailed analysis of PQG and GFE.
For these ablation studies, we used the same settings as mentioned in the main paper. Note that we used camera images as input and ResNet50 \cite{resnet} as the backbone. Training was conducted on 1/4 of the nuScenes training dataset for 24 epochs. Evaluation was performed on the entire validation set.

\begin{table}[h]
\begin{minipage}[t]{0.49\linewidth}
  \centering
  \centering
\caption{Ablation study of $N_s$ parameter in GFE } 
\begin{adjustbox}{width=0.8\textwidth}
    \begin{tabular}{c|cccc}
    \toprule[1.2pt]
     $N_s$ & $AP_{ped}$ & $AP_{divider}$ & $AP_{boundary}$ & $mAP$  \\ \toprule[0.7pt]
     10 & 51.3 & 55.6 & 58.3 & 55.1 \\
     20 & \textbf{52.9} & 55.5 & 58.5 & \textbf{55.6}  \\  
     30 & 50.5 & \textbf{56.0} & \textbf{58.7} & 55.1 \\ 
     40 & 50.7 & 55.8 & 57.9 & 54.8 \\
     \bottomrule[1.2pt]
    \end{tabular}
\end{adjustbox}
\label{ablation_sampling}
\end{minipage}
\hspace{1mm}
\begin{minipage}[t]{0.49\linewidth}
  \centering
  \centering
\caption{Ablation study of the {\it Max Survival} in GFE} 
\begin{adjustbox}{width=0.95\textwidth}
    \begin{tabular}{c|cccc}
    \toprule[1.2pt]
     {\it Max Survival} & $AP_{ped}$ & $AP_{divider}$ & $AP_{boundary}$ & $mAP$  \\ \toprule[0.7pt]     
      & 51.2 & \textbf{55.6} & 58.1 & 55.0 \\
     \checkmark & \textbf{52.9} & 55.5 & \textbf{58.5} & \textbf{55.6} \\ \bottomrule[1.2pt]
    \end{tabular}
\end{adjustbox}
\label{ablation_gridmax_gfe}
\end{minipage}%
\end{table}

\textbf{Effectiveness under Different $N_s$ of GFE.}
We investigated the impact of the parameter $N_s$, which represents the number of sampling points used in GFE.
Table \ref{ablation_sampling} shows the performance of Mask2Map evaluated for various values of $N_s=\{10, 20, 30, 40\}$. 
Mask2Map achieves the best performance when $N_s$ is 20, and degrades as $N_s$ becomes larger or smaller than these values. 
This is likely due to the fact that when the $N_s$ gets too large, a lot of redundant points can be used to extract geometric features.

\textbf{Effect of Max Survival in GFE.}
Table \ref{ablation_gridmax_gfe} demonstrates the effectiveness of the {\it Max Survival} employed in GFE, which is used to select the strongest pixel from the non-overlapping window of size $G \times G$ sliding on BEV Segmentation Masks. 
The performance of Mask2Map is enhanced by 0.6\% $mAP$ through the utilization of the {\it Max Survival} in GFE. 

\begin{table}[h]
\begin{minipage}[t]{0.49\linewidth}
  \centering
  \centering
\caption{Ablation study of $\tau_P$ parameter of PQG} 
\begin{adjustbox}{width=0.85\textwidth}
    \begin{tabular}{c|cccc}
    \toprule[1.2pt]
     $\tau_P$ & $AP_{ped}$ & $AP_{divider}$ & $AP_{boundary}$ & $mAP$  \\ \toprule[0.7pt]
     0.1 & 51.3 & 55.1 & 58.1 & 54.8 \\ 
     0.2 & 51.5 & 55.1 & 58.3 & 55.0 \\ 
     0.3 & \textbf{52.9} & \textbf{55.5} & \textbf{58.5} & \textbf{55.6}  \\  
     0.4 & 52.2 & 55.3 & 58.2 & 55.2 \\ 
     0.5 & 51.9 & 55.2 & 58.1 & 55.1 \\ 
     \bottomrule[1.2pt]
    \end{tabular}
\end{adjustbox}
\label{ablation_PFG_threshold}

\end{minipage}
\hspace{1mm}
\begin{minipage}[t]{0.49\linewidth}
  \centering
  
\centering
\caption{Ablation study under different $\tau_G$ parameter in GFE}
\begin{adjustbox}{width=0.8\textwidth}
    \begin{tabular}{c|cccc}
    \toprule[1.2pt]
     $\tau_G$ & $AP_{ped}$ & $AP_{divider}$ & $AP_{boundary}$ & $mAP$  \\ \toprule[0.7pt]
     0.6 & 51.1 & 55.2 & \textbf{59.0} & 55.1 \\ 
     0.7 & 51.3 & 55.5 & 58.9 & 55.3 \\  
     0.8 & \textbf{52.9} & 55.5 & 58.5 & \textbf{55.6}  \\  
     0.9 & 50.4 & \textbf{56.0} & 58.6 & 55.0 \\
     \bottomrule[1.2pt]
    \end{tabular}
\end{adjustbox}
\label{ablation_GFE_threshold}







\end{minipage}%
\end{table}

\textbf{Performance versus $\tau_P$ and $\tau_G$.}
Table \ref{ablation_PFG_threshold} and Table \ref{ablation_GFE_threshold} show ablation experiments for the thresholds $\tau_P$ and $\tau_G$, applied to filter the BEV Segmentation Masks in PQG and GFE, respectively.
$\tau_P$ is set to a lower threshold to capture comprehensive positional information, while $\tau_G$ is set to a higher threshold to extract accurate geometric information of map instances.
In our experiments, both $\tau_P$ and $\tau_G$ show the best performance in our experiments at 0.3 and 0.8, respectively.

\section{Extensive Qualitative Results}
In this section, we present qualitative results on the Argoverse2 and nuScenes benchmark datasets, including different scenarios, various weather conditions, failure cases, comparisons with various models, and initial predicted maps.

\begin{figure}[h]
    \centering
    \includegraphics[width=0.75\textwidth]{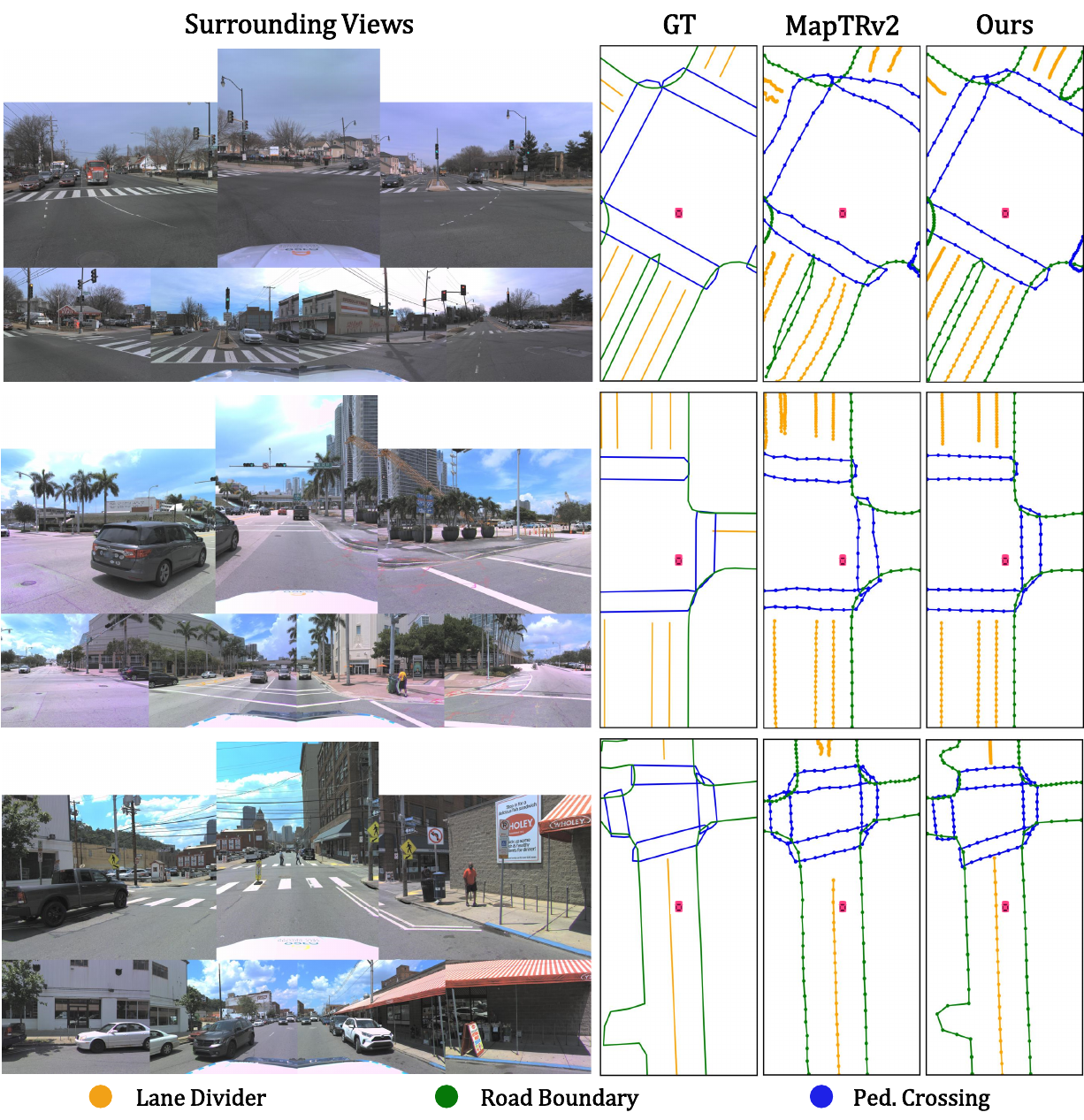}
    \caption{\textbf{Qualitative results under different scenarios on the Argoverse2 validation set.} We compared our method with MapTRv2. Both methods utilized multi-view camera images as input and employed ResNet50 \cite{resnet} as a backbone.
    }
    \label{sup_av2}
\end{figure}

\subsection{Qualitative Examples on Argoverse2}
We provide a qualitative comparison with the SOTA method \cite{MapTRv2} on the Argoverse2 validation set.
Figure \ref{sup_av2} shows enhancements in geometric detail compared to the result of MapTRv2 \cite{MapTRv2}.

\newpage
\pagebreak
\clearpage
\subsection{Additional visualization on nuScenes validation set.}

\begin{figure}[t]
    \centering
    \includegraphics[width=0.95\textwidth]{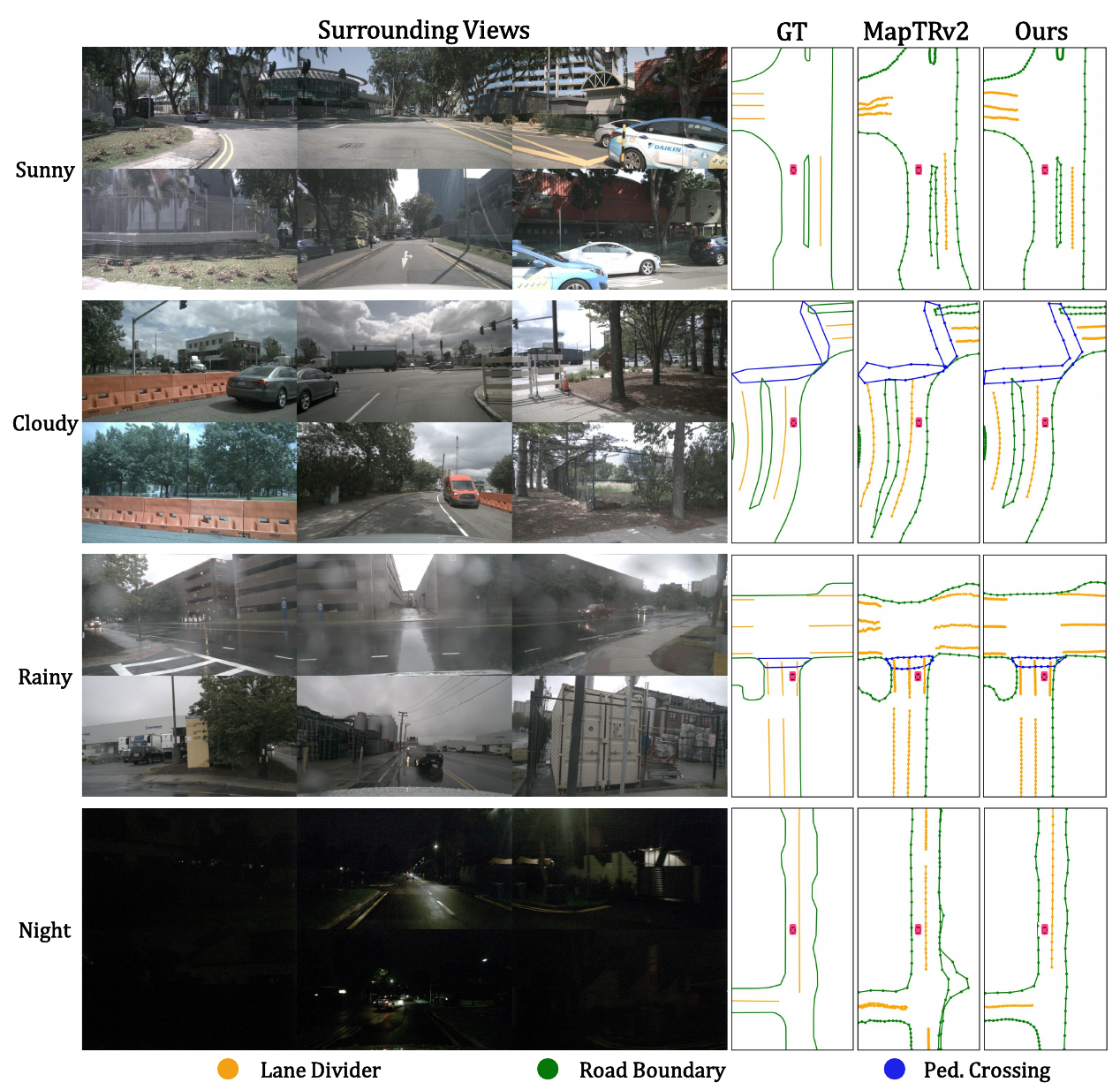}
    \caption{{\bf Qualitative results under various weather conditions on nuScenes validation set.} We used camera images as input and utilized ResNet50 \cite{resnet} as a backbone for comparisons.
    }    
    \label{sup_qualitative_plot}
\end{figure}

\textbf{Qualitative Results under Various Weather.}
We present additional qualitative comparisons with the SOTA method \cite{MapTRv2} under various weather conditions in Figure \ref{sup_qualitative_plot}. Even in more challenging conditions, such as rain or night, Mask2Map maintains stable and robust map construction quality compared to \cite{MapTRv2}.

\newpage

\begin{figure}[t]
    \centering
    \includegraphics[width=\textwidth]{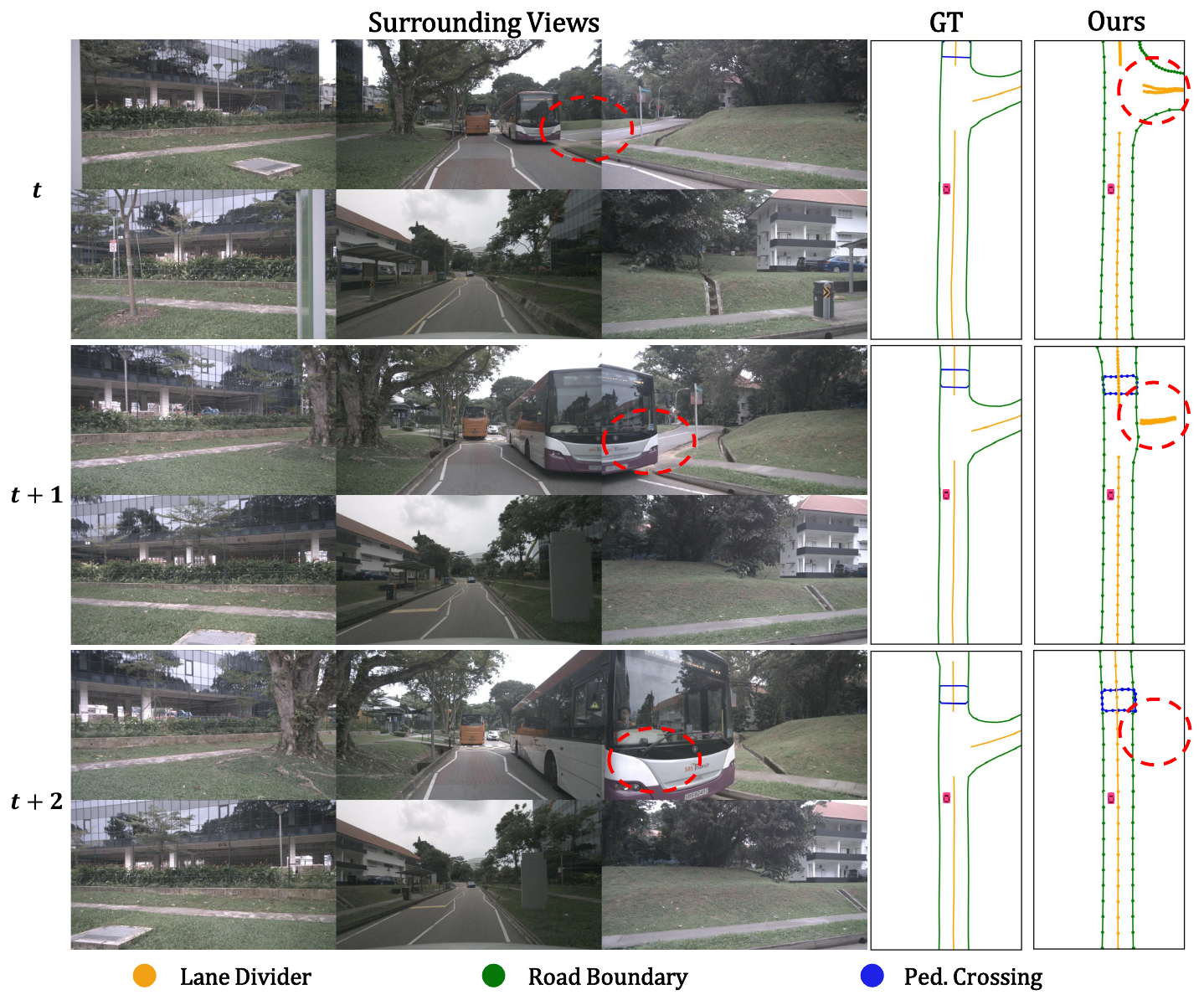}
    \caption{\textbf{Visualization of failure cases on nuScenes validation set.} 
    The regions marked by a red circle emphasize the failed map elements generated by our proposed model.
    The {\it $\textbf{t}$} denotes a time step, with {\it $\textbf{t}$}$+\textbf{1}$ and {\it $\textbf{t}$}$+\textbf{2}$ indicating subsequent steps.
    }
    \label{sup_failure_plot_1}
\end{figure}

\textbf{Failure Case Analysis.}
While Mask2Map significantly improves the performance of HD map construction, it still lacks the capability to accurately construct HD maps in scenarios where moving objects obstruct the view.
As shown in Figure \ref{sup_failure_plot_1}, we provide a visualization of the failure case. 
In the first row, Mask2Map constructs the map elements of the right-turn road marked with a red ellipse. 
However, in subsequent times, it fails to construct the map elements of the right-turn road due to occlusion caused by the movement of a bus. It is expected that utilizing temporal information before the occurrence of occlusion may improve these challenges.

\newpage

\begin{figure}[h!]
    \centering
    \includegraphics[width=0.90\textwidth]{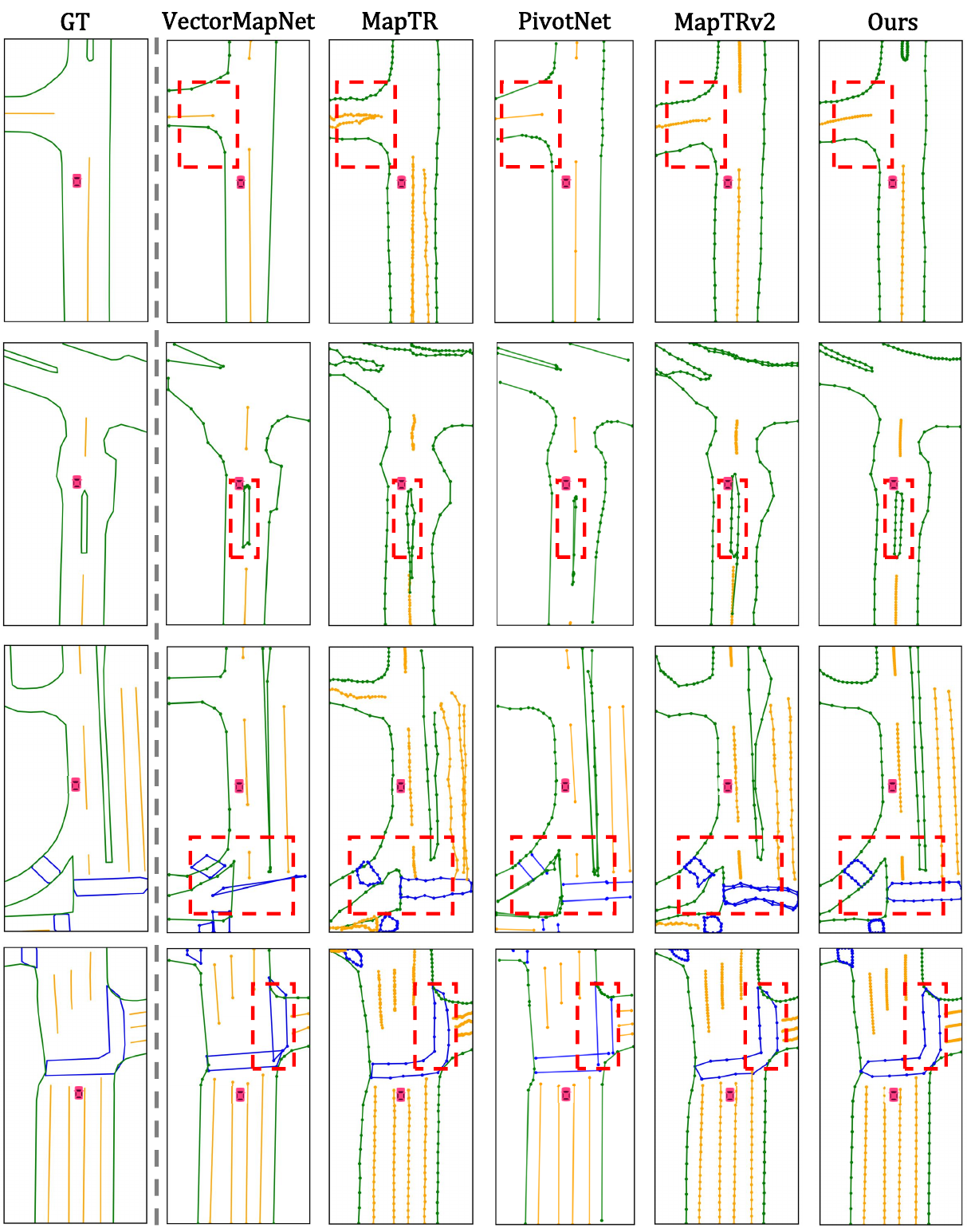}
    \caption{\textbf{Comparison with SOTA methods using qualitative visualization under different scenarios on the nuScenes validation set.} From the first to the sixth column: Ground truth, VectorMapNet \cite{VectorMapNet}, MapTR \cite{MapTR}, PivotNet \cite{PivotNet}, MapTRv2 \cite{MapTRv2}, and Mask2Map.
    }
    \label{sup_sota_methods}
\end{figure}

\textbf{Qualitative Results under Different Scenarios.}
We provide qualitative comparisons with SOTA methods under complex and various driving scenarios in Figure \ref{sup_sota_methods}. For a fair comparison, these methods employ the ResNet50 \cite{resnet} backbone and use multi-view camera images as input. 
Mask2Map constructs map shapes more smoothly than previous state-of-the-art models (see the red rectangle in Figure \ref{sup_sota_methods}), demonstrating robustness to various driving scenes. 

\newpage
\begin{figure}[!]
    \centering
    \includegraphics[width=1.0\textwidth]{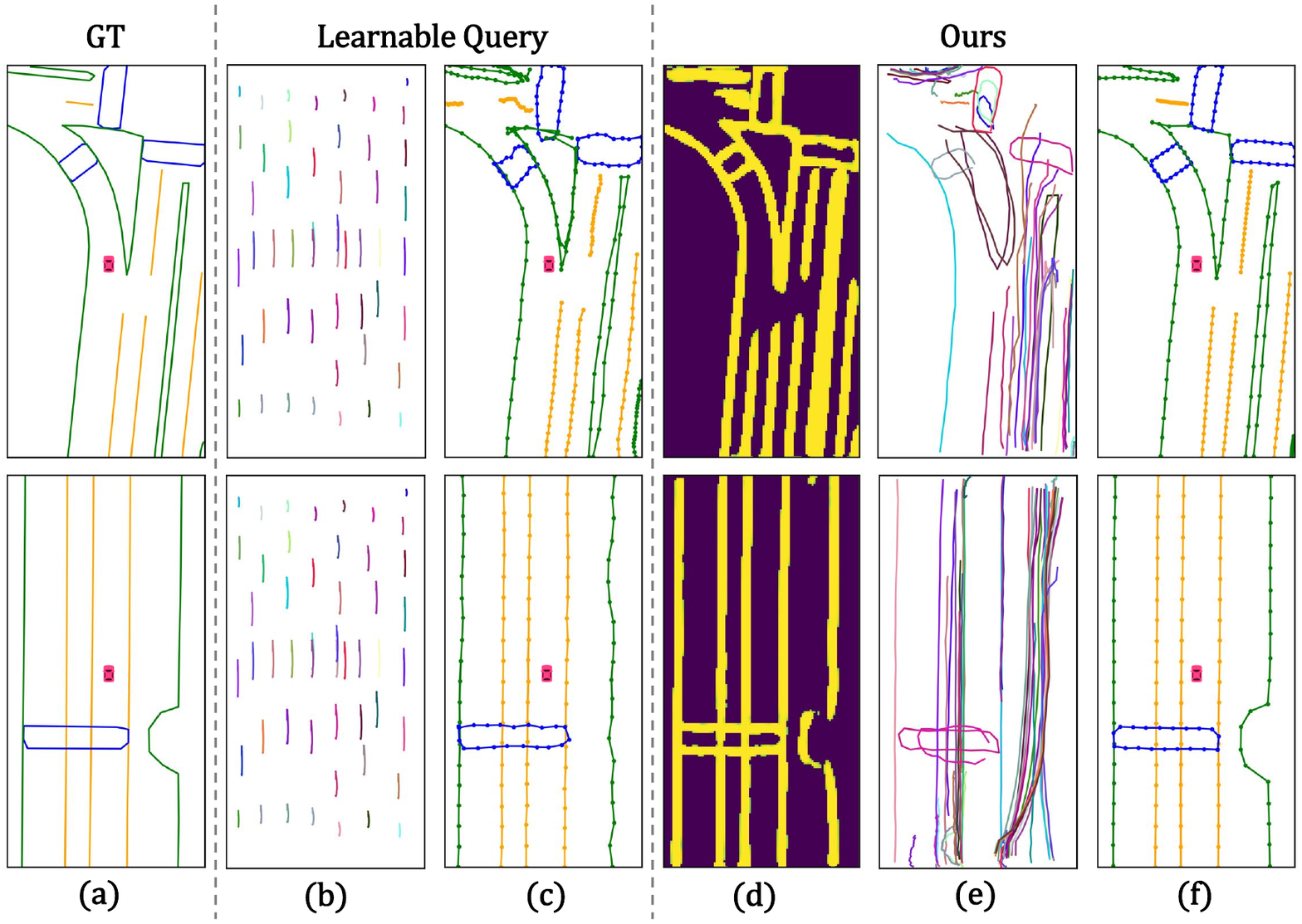}
    \caption{\textbf{Visualization of predicted outputs of MapTRv2 [\textcolor{green}{21}] and Mask2Map} : 
    (a) the ground truth, (b) and (c) results of MapTRv2; (b) maps predicted by learnable queries before decoding the Transformer decoder and (c) final predicted maps, (d), (e), and (f) outputs of Mask2Map; (d) BEV segmentation masks obtained from IMPNet, (e) maps predicted by Mask-Aware Hybrid Queries before decoding the Transformer decoder, and (f) final predicted maps.     
    }
    \label{qualitative_initial_ref}
\end{figure}

\textbf{Visualization of Initial Predicted Map.}

To demonstrate the effectiveness of Mask2Map, we represent a visualization of the initial map predicted by queries before decoding the Transformer decoders of MapTRv2 \cite{MapTRv2} and Mask2Map (Figure \ref{qualitative_initial_ref}), respectively. The initial predicted maps of MapTRv2 show identical results across different scenes (Figure \ref{qualitative_initial_ref} (b)). In contrast, the initial maps generated by Mask2Map show different results depending on the scene (Figure \ref{qualitative_initial_ref} (e)). This result demonstrates the effectiveness of using Mask-Aware Hybrid Queries that capture the semantic geometrical features of instances from a global perspective (Figure \ref{qualitative_initial_ref} (d)).

\end{document}